\documentclass[conference]{IEEEtran}

\usepackage{amsmath}

\usepackage{tikz}

\usepackage{tikz}
\usepackage{pgfplots}
\usepackage{array}
\pgfplotsset{compat=1.18}
\usepackage{wrapfig}
\usepackage{natbib}  
\usepackage{algorithm}
\usepackage{algpseudocode}

\usepackage{CJK}
\usepackage[utf8]{inputenc} 
\usepackage[T1]{fontenc}    
\usepackage{hyperref}       
\usepackage{url}            
\usepackage{booktabs}       
\usepackage{graphicx}
\usepackage{amsfonts}       
\usepackage{nicefrac}       
\usepackage{microtype}      
\usepackage{xcolor}         

\usepackage{multirow} 
\usepackage{tcolorbox}
\definecolor{primaryred}{HTML}{B01E3A}
\definecolor{primarybg}{HTML}{B01E3A}
\definecolor{xhsred}{HTML}{B01E3A}
\definecolor{xhsbg}{HTML}{FFFBFA}  
\usepackage{listings}     
\usepackage[utf8]{inputenc}
\usepackage{enumitem}
\usepackage{forest}
\usepackage{fancyvrb}

\newtcolorbox{finding}[1]{
  before={\par\noindent},
  colback=xhsbg!10,
  colframe=xhsred!70,
  title=Finding #1,
  fonttitle=\bfseries
}

\newcommand{\method}{\textsc{AgentGroupChat}-V2 }

\title{\method: Divide-and-Conquer Is What LLM-Based Multi-Agent System Need}

\author{
  \textbf{
  Zhouhong Gu$^{*, 1}$\thanks{}, 
  Xiaoxuan Zhu$^{*, 1}$, 
  Yin Cai$^2$,
  Hao Shen$^1$,
  Xingzhou Chen$^1$,}\\
  \textbf{
  Qingyi Wang$^1$,
  Jialin Li$^1$,
  Xiaoran Shi$^1$,
  Haoran Guo$^3$,
  Wenxuan Huang$^4$,}\\
  \textbf{
  Hongwei Feng$^1$,
  Yanghua Xiao$^1$,
  Zheyu Ye$^5$,
  Yao Hu$^5$, 
  Shaosheng Cao$^5$}\\
  $^1$Shanghai Key Laboratory of Data Science, School of Computer Science, Fudan University\\
  $^2$Fudan University, $^3$Rhine AI, $^4$East China Normal University,
  $^5$Xiaohongshu\\
  * Equal Contribution \\
  \{zhgu22, xxzhu22, haoshen22, xzchen24, xrshi21, wangqingyi22, jialingli22\}@m.fudan.edu.cn\\
  \{tanghuang1, guoba1\}@xiaohongshu.com, \{yincai, shawyh, hwfeng\}@fudan.edu.cn
}

\begin{document}

\maketitle
\begin{abstract}
Large language model based multi-agent systems have demonstrated significant potential in social simulation and complex task resolution domains. 
However, current frameworks face critical challenges in system architecture design, cross-domain generalizability, and performance guarantees, particularly as task complexity and number of agents increases. 
We introduces \method, a novel framework addressing these challenges through three core innovations: 
(1) a divide-and-conquer fully parallel architecture that decomposes user queries into hierarchical task forest structures enabling dependency management and distributed concurrent processing.
(2) an adaptive collaboration engine that dynamically selects heterogeneous LLM combinations and interaction modes based on task characteristics.
(3) agent organization optimization strategies combining divide-and-conquer approaches for efficient problem decomposition. 
Extensive experiments demonstrate \method's superior performance across diverse domains, achieving 91.50\% accuracy on GSM8K (exceeding the best baseline by 5.6 percentage points), 30.4\% accuracy on competition-level AIME (nearly doubling other methods), and 79.20\% pass@1 on HumanEval. 
Performance advantages become increasingly pronounced with higher task difficulty, particularly on Level 5 MATH problems where improvements exceed 11 percentage points compared to state-of-the-art baselines. These results confirm that \method provides a comprehensive solution for building efficient, general-purpose LLM multi-agent systems with significant advantages in complex reasoning scenarios.
Code is available at \url{https://github.com/MikeGu721/AgentGroupChat-V2}.
\end{abstract}


\begin{CJK}{UTF8}{gbsn}

\section{Introduction}
\begin{figure*}
    \centering
  \includegraphics[width=\textwidth]{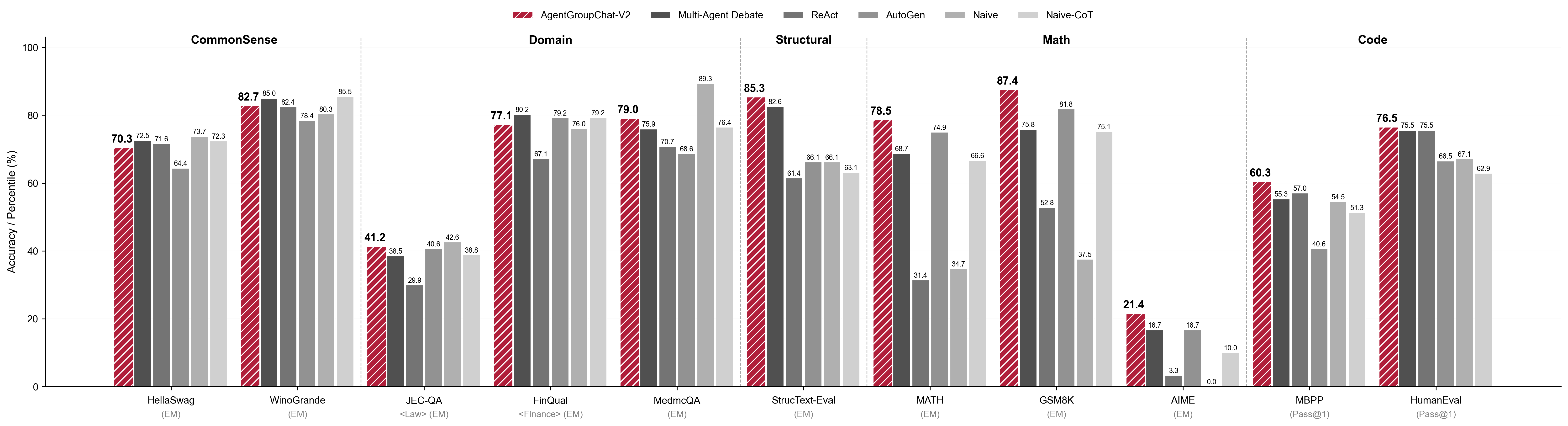}
  \includegraphics[width=\textwidth]{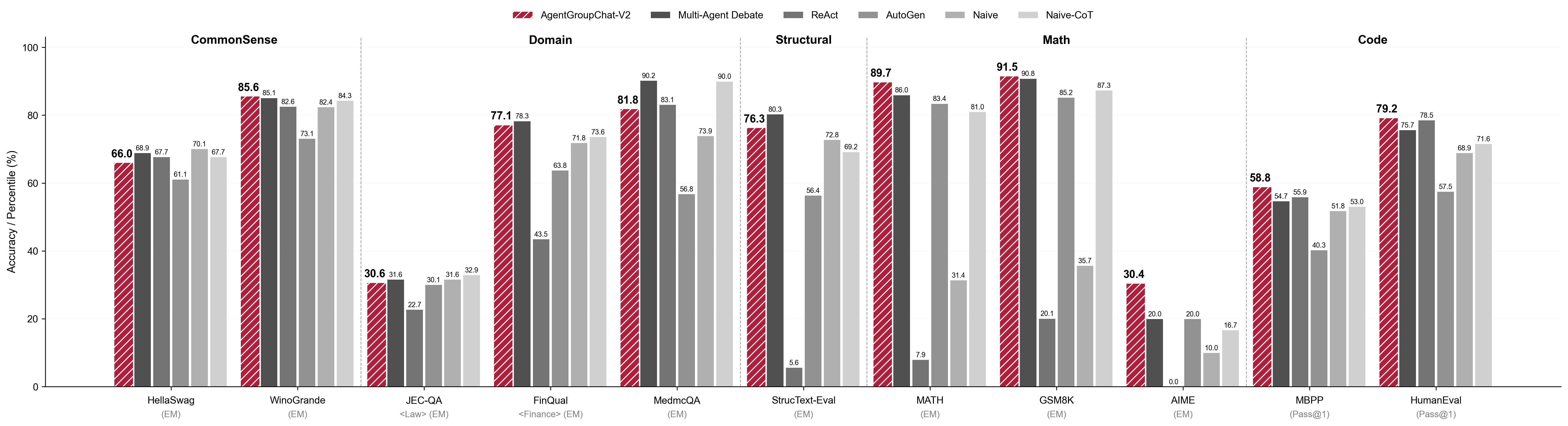}
\caption{
  \textbf{Upper:} The performance of Qwen2.5-72B.
  \textbf{Downer:} The performance of Llama3.1-70B.
  Both models are evaluated across five diverse reasoning domains: commonsense reasoning, domain-specific knowledge, structural text understanding, mathematics, and code generation.
  \method consistently outperforms existing multi-agent approaches and baseline methods across both models and all benchmark categories, demonstrating the effectiveness, robustness and promising of multi-agent approach.
}
\label{fig:teaser}
\end{figure*}

In recent years, research on LLM-based (large language model) multi-agent systems has grown substantially~\citep{guo2024large,li2023camel,xi2023rise,liang2024taskmatrix}. 
These systems demonstrate considerable potential in two primary application domains: social simulation~\citep{gao2024large, gao2023s3, gu2024agentgroupchat,park2023generative,zhou2023sotopia} and complex task resolution~\citep{hong2023metagpt, wu2023autogen,qian2024chatdev,xiong2023examining}.

In social simulation scenarios, these systems facilitate the modeling of human-like social interactions and behavioral patterns~\citep{park2023generative,bates1994role}, providing valuable insights for sociological research, policy development, and risk assessment~\citep{lee2023can,zhang2024electionsim}.
In task-oriented scenarios, collaborative agent networks demonstrate enhanced complex reasoning, planning~\citep{huang2024,wei2022chain}, and creative problem-solving capabilities through distributed intelligence~\citep{du2023improving}. 
Compared to traditional single-agent systems~\citep{chen2024travel,qin2023toolllm}, multi-agent architectures not only enable the integration of complementary knowledge~\citep{wooldridge2009introduction}, but also generate emergent problem-solving strategies through structured inter-agent interactions and hierarchical collaboration~\citep{ferber1999multi}, thereby achieving superior collective intelligence~\citep{dafoe2021cooperative}.

However, as the application scope of LLM multi-agent systems expands and task complexity increases, existing frameworks face systematic challenges~\citep{guo2024large,ouyang2022training,cemri2025multi}. 
First, existing solutions are predominantly designed for specific domains: systems like S3~\citep{gao2023s3} and AgentGroupChat~\citep{gu2024agentgroupchat} focus on social simulation scenarios, while MetaGPT~\citep{hong2023metagpt} and AutoGen~\citep{wu2023autogen} target specific tasks such as software development, lacking general-purpose design for handling diverse problem types~\citep{shoham2008multiagent}.
Second, multi-agent interactions often involve large-scale LLM calls and complex agent coordination, with existing systems commonly adopting sequential execution patterns, resulting in substantial computational resource waste and time overhead~\citep{Kumar2025,Zhu2025}. 
Finally, although multi-agent collaboration employs greater computational resources and should theoretically generally outperform single-agent approaches in terms of effectiveness, this is often not the case in practical scenarios~\citep{Zhang2023,Luo2025,cemri2025multi}.

The fundamental issue underlying the aforementioned challenges lies in the prevalent lack of effective collaboration schemes for general-purpose scenarios in existing LLM-based multi-agent systems. We argue that divide-and-conquer represents a key approach to addressing this fundamental problem. The divide-and-conquer strategy encompasses not only the decomposition and processing of complex tasks, breaking down large tasks into a series of manageable subtasks, but also the optimization and decomposition of multi-agent collaboration itself, enabling each agent to focus on specific aspects of the task. This approach can significantly enhance task processing effectiveness and parallel execution efficiency, while effectively improving the overall collaborative performance of multi-agent systems through specialized division of labor among agents, thereby providing a viable pathway for constructing general-purpose multi-agent collaboration schemes~\citep{shoham2008multiagent,ferber1999multi}.

To construct truly practical LLM multi-agent systems, three core challenges must be systematically addressed: (1) System architecture challenge: LLM multi-agent systems are inherently resource-intensive applications, yet current systems lack architectures specifically designed to support large-scale multi-agent collaboration~\citep{ferber1999multi,Kumar2025}; (2) Generalizability challenge: Existing systems primarily employ customized designs for specific scenarios, lacking cross-domain adaptability and adaptive problem understanding capabilities~\citep{shoham2008multiagent,gao2023s3}; (3) Performance guarantee challenge: Many multi-agent systems fail to demonstrate clear performance advantages over single-agent approaches in practical applications, sometimes performing worse due to system complexity~\citep{wooldridge2009introduction,Zhang2023}.

The core innovations of \method include: 
1) A fully parallel architecture with three coordinated manager modules that enable efficient resource control and high-throughput operation of LLM-based multi-agent systems, supporting distributed cluster deployment where manager modules coordinate global tasks on central servers while group managers handle parallel processes across multiple servers, ensuring distributed concurrent processing at all LLM invocation points and significantly improving system throughput and resource utilization; 
2) Task-level divide-and-conquer through dynamic task tree decomposition, where complex queries are hierarchically broken down into manageable subtasks with optimized dependency management and parallel execution; 
3) Execution-level divide-and-conquer through specialized agent role assignment, where different LLMs assume distinct roles and focus on specific aspects of problem-solving, enabling adaptive collaboration and cross-domain generalization. Experimental results demonstrate that \method achieves significant performance improvements while maintaining system generalizability, with particularly outstanding advantages on high-difficulty tasks (Figure~\ref{fig:teaser}), delivering substantial improvements across mathematical reasoning, code generation, and complex reasoning scenarios~\citep{qin2023toolllm,park2023generative,hong2023metagpt,wu2023autogen}.

\begin{figure*}[t]
    \centering
    \resizebox{\textwidth}{!}{\includegraphics{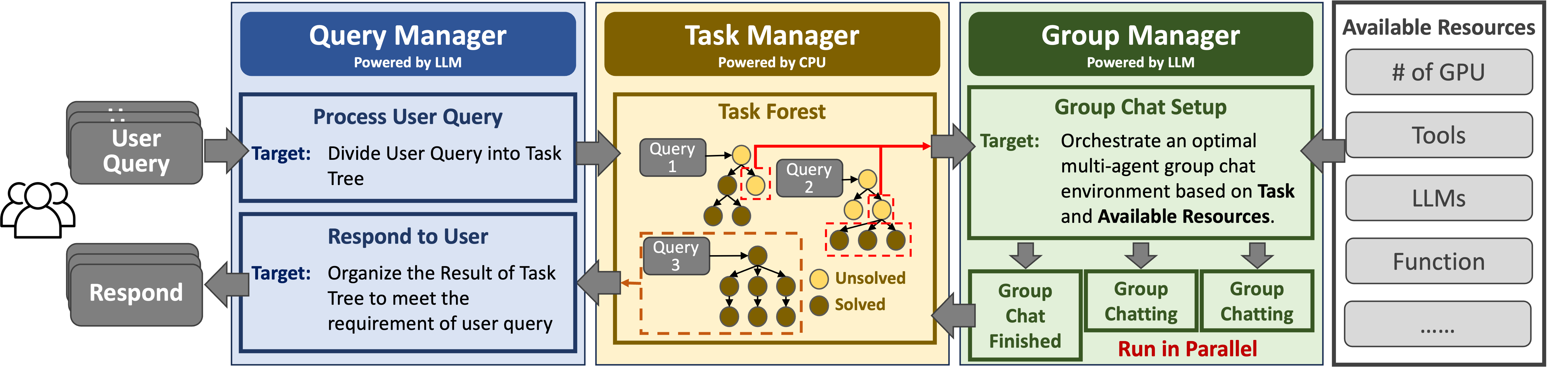}}
    \caption{Illustration of AgentGroupChat-V2 framework, which composes of three main components: Query Manager, Task Manager and Group Manager.
    The framework illustrates the complete workflow from user query processing through task decomposition and management to multi-agent group chat execution, with arrows indicating data flow between components.
    Task Forest visualization demonstrates how queries are transformed into hierarchical task structures with solved (brown) and unsolved (yellow) nodes, while parallel group chats are carrying out in Group Manager.}
    \label{fig:task_framework}
\end{figure*}

\section{Related Work}
LLM-based multi-agent systems have demonstrated transformative potential in two key domains~\cite{li2023camel,wooldridge2009introduction}:

\subsection{Social Dynamics Simulation}
Recent advances integrate cognitive architectures with behavioral economics principles~\cite{bates1994role}.Xie et al.~\cite{xie2024can} and Han et al.~\cite{han2023guinea} develop game-theoretic frameworks capturing trust dynamics and market competition through iterative belief-updating mechanisms~\cite{dafoe2021cooperative}.Sociological simulations leverage event-driven architectures, with Park et al.~\cite{park2023generative} modeling opinion evolution through social interaction cascades, while Zhang et al.~\cite{zhang2024electionsim} employ hierarchical Bayesian networks for electoral behavior prediction~\cite{lee2023can}.
Emerging platforms like Gu's group chat simulator~\cite{gu2024agentgroupchat} and Liu's rumor-control Twitter emulator~\cite{liu2024skepticism} demonstrate practical applications in digital social dynamics~\cite{gao2023s3}.

\subsection{Collaborative Problem-Solving}
Cutting-edge systems employ structured debate protocols and knowledge fusion mechanisms~\cite{du2023improving}.Xiong's FORD framework~\cite{xiong2023examining} enhances reasoning through tri-phase argumentation processes, complemented by Du's knowledge graph-based consensus formation~\cite{du2023improving}.Software engineering innovations like Qian's dialogue-driven development~\cite{qian2024chatdev} and Hong's documentation-centric workflow~\cite{hong2023metagpt} establish new paradigms for AI-assisted programming~\cite{wang2023voyager}.Domain-specific implementations showcase methodological cross-pollination, from Sun's legal argumentation system with adversarial validation~\cite{sun2024lawluo} to Yu's cognitive conflict-driven MOOC platform~\cite{yu2024mooc}, demonstrating versatile problem-solving architectures~\cite{ferber1999multi}.

\section{Framework of \method}

Building upon the AgentGroupChat framework \cite{gu2024agentgroupchat}, this research presents AgentGroupChat V2. As illustrated in Figure \ref{fig:task_framework}, \method employs a tripartite architecture comprising Query Manager, Task Manager, and Group Manager. This modular structure facilitates data exchange and function invocation through standardized interfaces and communication protocols. The architectural design of \method enables independent scaling of each module according to computational requirements, supporting horizontal expansion of the system's processing capabilities.

\subsection{Query Manager}

Query Manager functions as the frontend interaction component, integrating large language models as inference engines for direct user interaction. Its core functions include receiving User Queries, decomposing them into task tree structures through semantic understanding and task analysis, and forwarding these structures to the Task Manager for processing. Upon task completion, Query Manager receives the processing results, standardizes and integrates them, ensuring the response content meets user query requirements and is presented in an appropriate format.

\subsection{Task Manager}

Task Manager serves as the central coordination component, powered by CPU, implementing task flow management. A typical \method system is configured with a single Task Manager responsible for maintaining the state and execution of the entire task forest (containing multiple task trees derived from user queries). This module constructs a comprehensive task tracking system that records bidirectional relationships between tasks—child tasks link to their parent tasks, and parent tasks map to all their child tasks, forming a complete task dependency graph. Upon receiving tasks allocated by the Query Manager, the Task Manager determines task distribution strategies based on task characteristics and system load conditions. For hierarchically structured task groups, Task Manager implements information flow management, ensuring that processing results from child tasks serve as background knowledge for parent task processing workflows. For tasks without dependencies, Task Manager can distribute them to Group Manager instances based on resource availability, enabling parallel processing. When tasks are completed, Task Manager updates the task tree status, and when all tasks in a task tree have obtained results, the entire task tree is transmitted to Query Manager.

\subsection{Group Manager}

Group Manager functions as the execution component, responsible for organizing and coordinating multi-agent collaborative activities to accomplish specific tasks. This module incorporates large language models, with its primary function occurring during the group chat preparation phase, including selecting appropriate LLMs as inference engines for agents based on task requirements, allocating suitable workspace and object resources for each agent, and efficiently scheduling computational resources. Group Manager can be scaled according to system load, with multiple instances operating in parallel, each managing independent agent groups, enabling efficient utilization of system computational resources.

\section{Group Chat Design}
\begin{figure}[t] \centering \resizebox{0.5\textwidth}{!}{\includegraphics{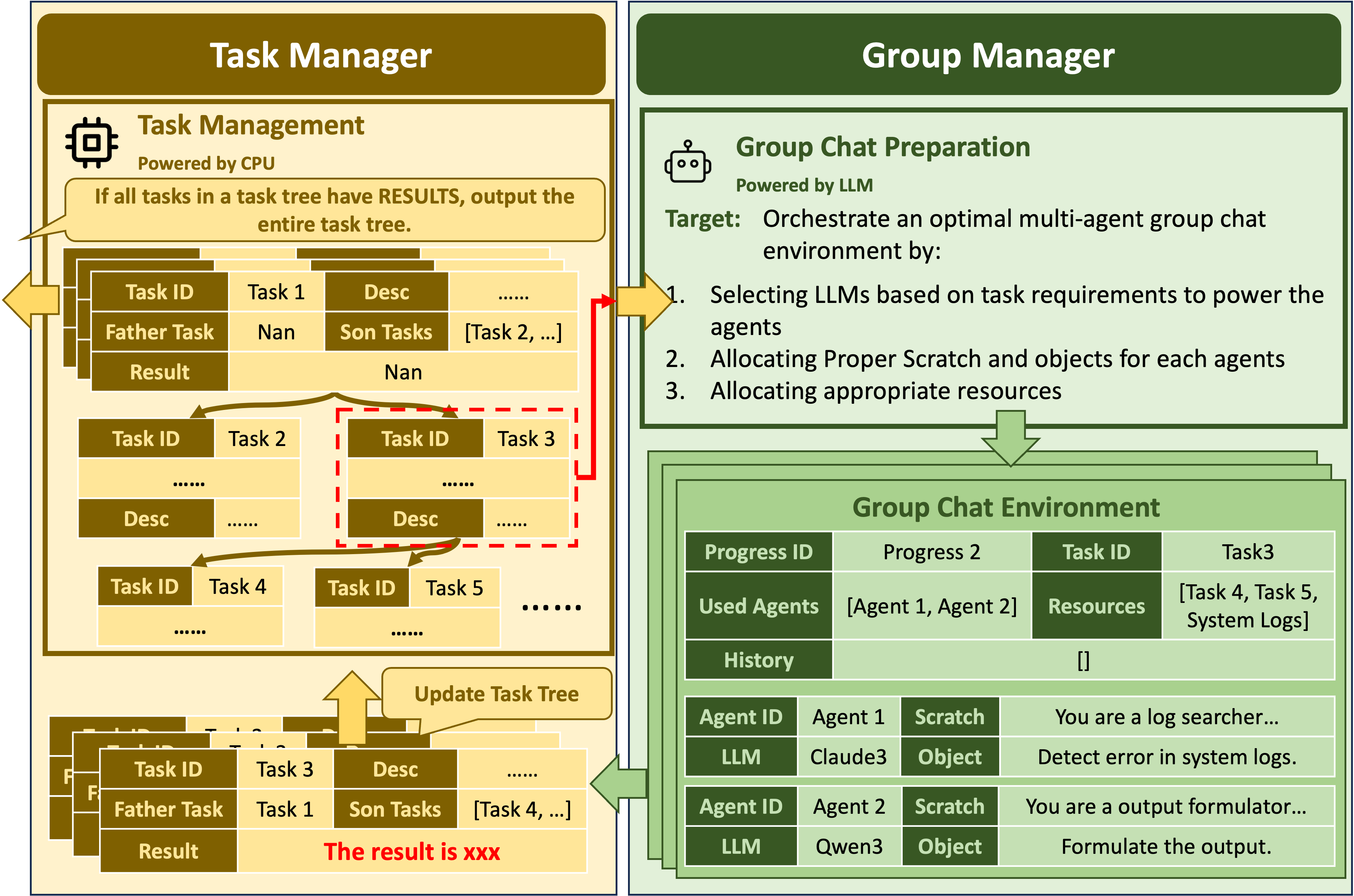}} \caption{Detailed implementation of Task Manager and Group Manager modules showing: (1) Task Management structure with hierarchical organization of tasks, including Task ID, descriptions, parent-child relationships, and result tracking; (2) Group Chat Preparation process outlining agent selection criteria and resource allocation strategies; and (3) Group Chat Environment configuration displaying the agent infrastructure with Claude3 and Qwen3 LLMs assigned specific roles, workspace allocations, and specialized functions for systematic problem-solving through collaborative interactions.} \label{fig:detail_framework} \end{figure}

This section elaborates on the organizational implementation mechanisms of group chat in the \method system. As the core element of multi-agent collaboration, the design of group chat directly impacts the system's problem-solving capabilities. As shown in Figure \ref{fig:detail_framework}, the system employs structured approaches to task organization and agent grouping to facilitate effective multi-agent problem solving.

\subsection{Task}

In the \method system, Task serves as the basic unit of task processing with a clearly structured definition. As shown in Figure \ref{fig:detail_framework}, each Task includes core attributes: Task ID (unique identifier), description information (Desc), parent task association (Father Task), child task collection (Son Tasks), and processing result (Result). When constructing task trees, the system marks the parent task of a root task as "Nan," while leaf tasks have empty child task collections. Tasks in the system follow a state transition process: initialization state (creating task structure without allocating execution resources) → waiting state (awaiting completion of prerequisite dependent tasks) → execution state (assigned to Group Manager for processing) → completion state (processing result generated) or failure state (errors encountered during processing).

\subsection{Group}

In the \method system, Group represents a multi-agent collaborative work unit created and managed by Group Manager. The Group Chat Environment shown in Figure \ref{fig:detail_framework} illustrates the core structure of a Group. Each Group contains basic information including progress identifier (Progress ID), participating agent list (Used Agents), task association (Task ID), and related resources (Resources). Each agent (Agent) within a Group possesses a clear identity (Agent ID), inference engine (LLM, typically large models such as Claude3 or Qwen3), dedicated workspace (Scratch), specific objects (Object), and a History field for storing interaction records. The Scratch space is primarily for the LLM's internal reference, while Object information is designed for human oversight, describing the agent's specific role and function (such as log searching or error detection). The History component stores the agent's conversation history, enabling persistent memory across multiple interaction rounds. Group Manager constructs complementary collaborative teams by assigning different functional positions and resource permissions to different agents. A Group's lifecycle includes: preparation phase (Group Manager selects and configures agents) → activity phase (agents engage in multi-turn dialogue to solve problems collectively) → result integration phase (transforming discussion outcomes into task outputs) → termination phase (releasing resources and updating system status).

\subsection{Group Environment Configuration}

Group environment configuration forms the foundational condition for effective group chat. Each group environment contains a unique Progress ID to identify the current progress, an associated Task ID indicating which task the group is responsible for processing, Used Agents recording the set of participating agents and their speaking order in the group chat, and a Resources field defining available system resources shared across the group, such as execution records of specific tasks or system logs.

The group environment configuration process is led by Group Manager. Based on task requirements transmitted by Task Manager, Group Manager makes intelligent decisions through its built-in large language model, selecting the most suitable inference engine for each agent according to available LLM types and their relevance to the task. This design allows the system to flexibly configure heterogeneous LLM combinations based on different task characteristics, while also allowing different identities to be assigned to LLMs, enabling each Agent to focus on a specific objective. Group Manager configures three key elements for each agent: Scratch space for granting the LLM specific identity cognition; Object information for recording the agent's specific tasks, providing task transparency to users; and History for storing records of conversations and insights gained during the interaction process.

\subsection{Group Chat Orchestration}

The group chat orchestration defines the complete lifecycle of collaborative problem-solving from initiation to conclusion. As shown in Algorithm \ref{algrm1}, group chat is coordinated by Group Manager, following a preset maximum number of action turns.

\begin{algorithm}[t]
\caption{StartGroupChat (GroupManager)}
\label{algrm1}
\small
\begin{algorithmic}[1]
\State \textbf{Input:} max\_action\_turn, agent\_ids, initial\_env
\State \textbf{Output:} final\_env, task\_result
\State env $\gets$ initial\_env
\For{turn = 1 to max\_action\_turn}
    \For{current\_agent in agent\_ids}
        \State env $\gets$ current\_agent.perceive(env)
        \State (message, target\_agent) $\gets$ current\_agent.decide\_action(env)
        \State dialogue\_history $\gets$ \Call{ExecuteAction}{current\_agent, env, message, target\_agent}
        \State env $\gets$ UpdateEnvironment(dialogue\_history, env)
    \EndFor
    \State discussion\_summary $\gets$ SummarizeGroupMessages(env)
    \State env $\gets$ UpdateEnvironment(discussion\_summary, env)
    \State is\_complete $\gets$ TaskManager.CheckTaskCompletion(env)
    \If{is\_complete}
        \State \Return env, ExtractTaskResult(env)
    \EndIf
\EndFor
\State \Return env, ExtractTaskResult(env)
\end{algorithmic}
\end{algorithm}

Algorithm \ref{algrm1} demonstrates the core process of group chat. The system receives the maximum number of action turns, a list of participating agents, and the initial environment state as inputs. In each action turn, agents sequentially perform the complete process of perceiving the environment, making decisions, executing interactions, and updating the environment. At the end of each round, the system generates a dialogue summary and checks for task completion, returning results early if the task is completed. This structured process ensures orderly multi-agent collaboration and goal orientation, while providing a flexible task termination mechanism.

\subsection{Agent Interaction}

Agent interaction is the core element of group chat. Algorithm \ref{algrm2} describes the execution process of a single agent action, supporting both broadcast messages and directed communication modes.

\begin{algorithm}[t]
\caption{\textsc{ExecuteAction}\{current\_agent, env, message, target\_agent\}}
\label{algrm2}
\small
\begin{algorithmic}[1]
\State \textbf{Input:} current\_agent, env, message, target\_agent, max\_chat\_turn
\State \textbf{Output:} dialogue\_history
    \State turn\_count $\gets$ 1
    \State dialogue\_history $\gets$ [ ]
    \If{target\_agent is AllGroupMembers}
    \State message $\gets$ GenerateMessage(current\_agent, env, message, target\_agent)
    \State Append (current\_agent, target\_agent, message) to dialogue\_history
    \State \Return dialogue\_history
    \EndIf
    \State sender $\gets$ current\_agent
    \State receiver $\gets$ target\_agent
    \While{turn\_count < 2 $\cdot$ max\_chat\_turn}
        \State response $\gets$ GenerateResponse(receiver, dialogue\_history, env)
        \If{not response}
            \State \Return dialogue\_history
        \EndIf
        \State Append (receiver, sender, response) to dialogue\_history
        \State temp $\gets$ sender
        \State sender $\gets$ receiver
        \State receiver $\gets$ temp
        \State turn\_count $\gets$ turn\_count + 1
    \EndWhile
    \State \Return dialogue\_history
\end{algorithmic}
\end{algorithm}

Algorithm \ref{algrm2} implements flexible interaction modes between agents. This algorithm receives the initiating agent, current environment, message type, and target agent as inputs. The initiating agent first generates an initial message; if the recipient is all members, it directly returns the broadcast message record; otherwise, it enters a dual-agent alternating dialogue loop, which continues until the maximum number of turns is reached or the receiving party decides to terminate the dialogue. This mechanism allows agents to flexibly select interaction methods and objects based on task requirements, achieving multi-level collaboration from broad information sharing to in-depth professional discussion.

\subsection{Chat Results Processing}

Chat results processing is a key link in ensuring effective transformation of multi-agent collaboration outcomes. Group Manager regularly summarizes dialogue content, extracting key information and phase conclusions, which are stored in agents' Memory to provide context support for subsequent interactions. The system simultaneously conducts quality assessment, examining whether the discussion has reached effective conclusions and whether it meets task requirements. Finally, the system performs format standardization processing on confirmed effective group chat results, forming structured task output to ensure compliance with Task Manager processing requirements.

\section{Experiment Setup}
\subsection{Task \& Benchmark}

\subsubsection{Mathematical Reasoning} 
The experimental evaluation encompasses three established mathematical reasoning benchmarks: 
\textbf{GSM8K} comprises 1,300 test instances of grade school mathematics problems.
\textbf{MATH} features 12,000 challenging problems spanning five difficulty levels, covering high school mathematics and competition-level mathematics. 
\textbf{AIME} collects American Invitational Mathematics Examination problems, featuring high-complexity mathematical problems from the prestigious competition, containing problems that assess advanced mathematical reasoning through exact match evaluation of numerical answers.
All benchmarks employ symbolic equivalence verification for accuracy assessment.

\subsubsection{Code Generation} 
The code generation capabilities are evaluated using two standardized benchmarks:
\textbf{MBPP} (Mostly Basic Python Problems) consists of 500 Python programming tasks accompanied by test cases, with performance assessed via unit-test pass rate.
\textbf{HumanEval} provides 164 function-level programming challenges, evaluated through the pass@k metric:
\begin{equation}
\text{Pass}@k = 1 - \frac{\binom{n-c}{k}}{\binom{n}{k}}
\end{equation}
where $n$ denotes the total number of generated solutions and $c$ represents correct solutions. We generate $n=10$ candidate solutions per problem and compute the average pass rates at $k=1/3/5$. 

\subsubsection{Domain-Specific Tasks} 
\textbf{FinQual} offers 1,000 multiple-choice and numerical calculation questions sampled from the CFA examinations and FinQA test sets, assessing financial reasoning and computational accuracy.
\textbf{JEC-QA} contains 26,365 multiple-choice questions sourced from Chinese legal professional qualification exam preparation materials, evaluating legal concept understanding and scenario analysis.
\textbf{MedmcQA} provides 194K multiple-choice questions from Indian medical entrance examinations, evaluating healthcare concepts and clinical reasoning.

\subsubsection{Structural Text Understanding} 
\textbf{StrucText-Eval} assesses models' capabilities in understanding and processing structured textual data with varying complexity levels.

\subsubsection{Commonsense Reasoning} 
\textbf{HellaSwag} offers 70,000 commonsense question-answering pairs formulated as contextual multiple-choice problems.
\textbf{WinoGrande} provides 44,000 pronoun resolution challenges presented in a binary selection format, requiring disambiguation through contextual reasoning.

\subsection{Baseline Methods}
We evaluate five representative baseline methods:

\textbf{Naive}: Direct task forwarding to a single LLM without prompt engineering or problem decomposition.
\textbf{Naive-CoT}~\cite{wei2022chain}: Single-agent paradigm employing chain-of-thought prompting to elicit explicit reasoning traces.
\textbf{ReAct}~\cite{yao2022react}: Structured single-agent framework decomposing problems through iterative Reasoning, Action, and Observation cycles.
\textbf{AutoGen}~\cite{wu2023autogen}: Programmable multi-agent conversation framework with AssistantAgent and UserProxyAgent components.
\textbf{Multi-Agent Debate}~\cite{liang2023encouraging}: Collaborative paradigm where multiple agents analyze problems through predefined dialogue sequences until consensus.

\subsection{Large Language Models}
We employ two state-of-the-art LLMs: \textbf{Qwen2.5-72B-Instruct} and \textbf{Llama-3.1-70B-Instruct-Turbo}, selected for their superior performance in general-purpose task handling and computational efficiency.

\section{Experiment}

\begin{table*}[t]
\centering
\small
\caption{Experiment result on Math Problem.}
\label{tab:math}
\begin{tabular}{cccccccccc}
\toprule
\multirow{2}{*}{\textbf{Baseline}} & \multirow{2}{*}{\textbf{LLM}} & \multirow{2}{*}{\textbf{GSM8K}} & \multicolumn{5}{c}{\textbf{MATH}} & \textbf{AIME}\\
\cmidrule(lr){4-8}
& & & \textbf{Level1} & \textbf{Level2} & \textbf{Level3} & \textbf{Level4} & \textbf{Level5} & \textbf{(2024)}\\
\midrule
\multirow{2}{*}{\textbf{Naive}} & Qwen2.5-72B & 37.52 & 60.17 & 49.41 & 37.16 & 26.49 & 21.78 & 0.0 \\
\cmidrule(lr){2-9}
& Llama-3.1-70B & 35.70 & 50.44 & 44.11 & 30.97 & 28.20 & 19.84 & 10.0\\
\midrule
\multirow{2}{*}{\textbf{Naive-CoT}} & Qwen2.5-72B & 75.13 & 85.84 & 77.65 & 75.22 & 63.68 & 48.25 & 10.0 \\
\cmidrule(lr){2-9}
& Llama-3.1-70B & 87.33 & 91.15 & 88.82 & 87.17 & 80.77 & 67.32 & 16.7 \\
\midrule
\multirow{2}{*}{\textbf{ReAct}} & Qwen2.5-72B & 52.76 & 53.10 & 35.88 & 36.73 & 28.21 & 19.46 & 3.3 \\
\cmidrule(lr){2-9}
& Llama-3.1-70B & 20.09 & 28.32 & 11.76 & 7.52 & 4.27 & 2.33 & 0.0 \\
\midrule
\multirow{2}{*}{\textbf{AutoGen}} & Qwen2.5-72B & \underline{81.80} & \underline{89.38} & \underline{85.29} & \textbf{88.05} & \textbf{71.79} & \underline{54.86} & \underline{16.7} \\
\cmidrule(lr){2-9}
& Llama-3.1-70B & 85.21 & 96.46 & \underline{92.94} & \underline{91.15} & \underline{82.48} & 66.93 & \underline{20.0} \\
\midrule
\multirow{2}{*}{\textbf{Multi-Agent Debate}} & Qwen2.5-72B & 75.81 & 85.84 & 79.41 & 75.66 & \underline{66.24} & 52.14 & \underline{16.7} \\
\cmidrule(lr){2-9}
& Llama-3.1-70B & \underline{90.82} & \underline{97.35} & \underline{92.94} & \textbf{92.48} & \textbf{85.90} & \underline{71.98} & \underline{20.0} \\
\midrule
\multirow{2}{*}{\textbf{\method}} & Qwen2.5-72B & \textbf{87.41} & \textbf{92.92} & \textbf{90.00} & \underline{84.07} & \textbf{71.79} & \textbf{59.10} & \textbf{21.4} \\
\cmidrule(lr){2-9}
& Llama-3.1-70B & \textbf{91.50} & \textbf{98.23} & \textbf{94.12} & 88.94 & 81.20 & \textbf{83.54} & \textbf{30.4} \\
\bottomrule
\end{tabular}
\end{table*}
\begin{table*}[t]
\centering
\small
\caption{Experiment result on Code Generation.}
\label{tab:code}
    \begin{tabular}{cccccccc}
        \toprule
        \multirow{2}{*}{\textbf{Baseline}} & \multirow{2}{*}{\textbf{LLM}} & \multicolumn{3}{c}{\textbf{MBPP}} & \multicolumn{3}{c}{\textbf{HumanEval}} \\
        \cmidrule(lr){3-5} \cmidrule(lr){6-8}
        & & \textbf{pass@1} & \textbf{pass@3} & \textbf{pass@5} & \textbf{pass@1} & \textbf{pass@3} & \textbf{pass@5} \\
        \midrule
        \multirow{2}{*}{\textbf{Naive}} & Qwen2.5-72B & 54.46 & 60.09 & 62.02 & 67.07 & 78.34 & 81.60\\
        \cmidrule(lr){2-8}
        & Llama-3.1-70B & 51.80 & 60.82 & 63.65 & 68.90 & 80.92 & 83.52 \\
        \midrule
        \multirow{2}{*}{\textbf{Naive-CoT}} & Qwen2.5-72B & 51.30 & 61.52 & 65.08 & 62.86 & 78.60 & 82.43 \\
        \cmidrule(lr){2-8}
        & Llama-3.1-70B & 53.04 & 60.51 & 62.75 & 71.58 & 81.23 & 83.56 \\
        \midrule
        \multirow{2}{*}{\textbf{ReAct}} & Qwen2.5-72B & \underline{57.02} & \textbf{65.34} & \textbf{68.50} & \underline{75.54} & \textbf{84.39} & \textbf{86.84} \\
        \cmidrule(lr){2-8}
        & Llama-3.1-70B & \underline{55.88} & \underline{62.39} & \underline{64.84} & \underline{78.53} & \textbf{85.71} & \textbf{87.66} \\
        \midrule
        \multirow{2}{*}{\textbf{AutoGen}} & Qwen2.5-72B & 40.60 & 54.33 & 59.27 & 66.46 & 80.83 & 84.22 \\
        \cmidrule(lr){2-8}
        & Llama-3.1-70B & 40.26 & 55.05 & 60.07 & 57.50 & 75.20 & 80.26 \\
        \midrule
        \multirow{2}{*}{\textbf{Multi-Agent Debate}} & Qwen2.5-72B & 55.28 & \underline{63.76} & \underline{66.67} & \underline{75.54} & \underline{83.19} & \underline{85.41} \\
        \cmidrule(lr){2-8}
        & Llama-3.1-70B & 54.68 & \textbf{63.02} & \textbf{65.75} & 75.67 & \underline{84.81} & \underline{87.65} \\
        \midrule
        \multirow{2}{*}{\textbf{\method}} & Qwen2.5-72B & \textbf{60.34} & \underline{63.76} & 64.45 & \textbf{76.46} & 82.31 & 84.15 \\
        \cmidrule(lr){2-8}
        & Llama-3.1-70B & \textbf{58.84} & 60.35 & 60.84 & \textbf{79.20} & 80.38 & 80.91 \\
        \bottomrule
    \end{tabular}
\centering
\end{table*}
\begin{table}[t]
\centering
\small
\caption{Experiment result on Commonsense Reasoning.}
\label{tab:commonsense}
\renewcommand{\arraystretch}{1.3}   
\setlength{\tabcolsep}{0.3pt}        
\begin{tabular}{ccccc}
\toprule
\textbf{Baselines} & \textbf{LLM} & \textbf{HellaSwag} & \textbf{WinoGrande} \\
\midrule
\multirow{2}{*}{\textbf{Naive}} 
& Qwen2.5-72B & \textbf{73.7} & 80.3 \\
\cmidrule(lr){2-4}
& Llama-3.1-70B & \textbf{70.1} & 82.4 \\
\midrule
\multirow{2}{*}{\textbf{Naive-CoT}}
& Qwen2.5-72B & 72.3 & \textbf{85.5} \\
\cmidrule(lr){2-4}
& Llama-3.1-70B & 67.7 & 84.3 \\
\midrule
\multirow{2}{*}{\textbf{ReAct}}
& Qwen2.5-72B & 71.6 & 82.4 \\
\cmidrule(lr){2-4}
& Llama-3.1-70B & 67.7 & 82.6 \\
\midrule
\multirow{2}{*}{\textbf{AutoGen}}
& Qwen2.5-72B & 64.4 & 78.4 \\
\cmidrule(lr){2-4}
& Llama-3.1-70B & 61.1 & 73.1 \\
\midrule
\multirow{2}{*}{\textbf{Multi-Agent Debate}}
& Qwen2.5-72B & \underline{72.5} & \underline{85.0} \\
\cmidrule(lr){2-4}
& Llama-3.1-70B & \underline{68.9} & \underline{85.1} \\
\midrule
\multirow{2}{*}{\textbf{\method}}
& Qwen2.5-72B & 70.3 & 82.7 \\
\cmidrule(lr){2-4}
& Llama-3.1-70B & 66.0 & \textbf{85.6} \\
\bottomrule
\end{tabular}
\end{table}

\begin{table*}[t]
\centering
\small
\caption{EM Accuracy for StrucText-Eval.}
\label{tab:struc}
\renewcommand{\arraystretch}{1.3}   
\setlength{\tabcolsep}{0.3pt}        
\begin{tabular}{ccccccccccc}
\toprule
\multirow{2}{*}{\textbf{Method}} & \multirow{2}{*}{\textbf{Model}} 
  & \multicolumn{3}{c}{\textbf{Width=1}} 
  & \multicolumn{3}{c}{\textbf{Width=2}}
  & \multicolumn{3}{c}{\textbf{Width=3}} \\
\cmidrule(r){3-5} \cmidrule(lr){6-8} \cmidrule(l){9-11}
  & & Depth=1 & Depth=2 & Depth=3 & Depth=1 & Depth=2 & Depth=3 & Depth=1 & Depth=2 & Depth=3 \\ 
\midrule
\multirow{2}{*}{\textbf{Naive}} 
 & Llama3.1-70B & 78.5 & 77.2 & 72.8 & 69.3 & 63.5 & \underline{54.2} & 66.8 & 53.3 & 42.7 \\
 & Qwen2.5-72B  & \underline{81.9} & \underline{80.7} & \underline{77.9} & \underline{72.5} & \underline{68.0} & \underline{55.4} & \underline{70.3} & \underline{56.8} & \underline{45.0} \\
\midrule
\multirow{2}{*}{\textbf{Naive-CoT}} 
 & Llama3.1-70B & \textbf{84.1} & \textbf{83.8} & \underline{76.2} & \textbf{75.5} & \textbf{70.3} & 53.7 & \textbf{73.2} & \underline{57.1} & \underline{43.5} \\
 & Qwen2.5-72B  & \textbf{86.7} & \textbf{85.3} & \textbf{78.3} & \textbf{77.8} & \textbf{71.4} & 54.8 & \textbf{75.5} & 55.2 & 42.3 \\
\midrule
\multirow{2}{*}{\textbf{ReAct}} 
 & Llama3.1-70B & 5.2 & 3.8 & 2.1 & 4.1 & 2.9 & 1.3 & 3.5 & 1.8 & 0.5 \\
 & Qwen2.5-72B  & 8.1 & 6.7 & 4.9 & 6.3 & 4.2 & 2.8 & 5.1 & 3.1 & 1.2 \\
\midrule
\multirow{2}{*}{\textbf{AutoGen}} 
 & Llama3.1-70B & 28.4 & 24.7 & 19.8 & 22.1 & 17.3 & 12.6 & 18.9 & 13.2 & 8.7 \\
 & Qwen2.5-72B  & 34.2 & 31.1 & 26.5 & 27.8 & 22.9 & 17.4 & 24.3 & 18.1 & 12.8 \\
\midrule
\multirow{2}{*}{\textbf{Multi-Agent Debate}} 
 & Llama3.1-70B & \underline{83.3} & \underline{82.2} & 74.8 & \underline{74.1} & \underline{68.9} & 52.2 & \underline{72.4} & 54.7 & 40.3 \\
 & Qwen2.5-72B  & 82.1 & 81.5 & 75.2 & 76.3 & 67.6 & 53.8 & 71.2 & 55.4 & 41.7 \\
\midrule
\multirow{2}{*}{\textbf{\method}} 
 & Llama3.1-70B & 81.7 & 80.8 & \textbf{79.1} & 73.6 & 69.2 & \textbf{58.4} & 72.8 & \textbf{62.9} & \textbf{48.7} \\
 & Qwen2.5-72B  & 83.5 & 82.9 & 77.3 & 76.7 & 70.8 & \textbf{59.2} & 73.6 & \textbf{64.7} & \textbf{52.1} \\
\bottomrule
\end{tabular}
\end{table*}

\begin{table*}[t]
\centering
\small
\caption{Experiment results on Law, Finance, and Medical multiple-choice QA tasks.}
\label{tab:domain}
    \begin{tabular}{ccccccc}
        \toprule
        \multirow{2}{*}{\textbf{Method}} & \multirow{2}{*}{\textbf{Model}} & \textbf{JEC-QA~(Law)} & \textbf{FinQual~(Finance)} & \textbf{MedmcQA~(Medical)} \\
        && \textbf{EM Accuracy} & \textbf{EM Accuracy} & \textbf{EM Accuracy} \\
        \midrule
        \multirow{2}{*}{\textbf{Naive}} & Llama-3.1-70B & 31.58 & 71.83 & 73.90 \\
        \cmidrule(lr){2-5}
        & Qwen2.5-72B & 42.56 & 76.00 & 89.30 \\
        \midrule
        \multirow{2}{*}{\textbf{Naive-CoT}} & Llama-3.1-70B & 32.93 & 73.60 & 89.97 \\
        \cmidrule(lr){2-5}
        & Qwen2.5-72B & 38.80 & 79.20 & 76.40 \\
        \midrule
        \multirow{2}{*}{\textbf{ReAct}} & Llama-3.1-70B & 22.71 & 43.50 & 83.10 \\
        \cmidrule(lr){2-5}
        & Qwen2.5-72B & 29.92 & 67.10 & 70.70 \\
        \midrule
        \multirow{2}{*}{\textbf{AutoGen}} & Llama-3.1-70B & 30.08 & 63.76 & 56.79 \\
        \cmidrule(lr){2-5}
        & Qwen2.5-72B & 40.60 & 79.16 & 68.60 \\
        \midrule
        \multirow{2}{*}{\textbf{Multi-Agent Debate}} & Llama-3.1-70B & 31.58 & 78.28 & 90.20 \\
        \cmidrule(lr){2-5}
        & Qwen2.5-72B & 38.50 & 80.20 & 75.90 \\
        \midrule
        \multirow{2}{*}{\textbf{\method}} & Llama-3.1-70B & 30.62 & 77.09 & 81.82 \\
        \cmidrule(lr){2-5}
        & Qwen2.5-72B & 41.20 & 77.11 & 79.00 \\
        \bottomrule
    \end{tabular}
\end{table*}

\begin{table*}[t]
\centering
\small
\caption{Role configuration for multi-agent mathematical problem-solving collaboration.}
\label{tab:abla_role}
\begin{tabular}{p{3.5cm}p{12cm}}
\toprule
\textbf{Role Type} & \textbf{Role Setting} \\
\midrule
\multirow{1}{*}{General Role} & 
Agent-001 is a \textcolor{xhsred}{math expert}.\\
\midrule
\multirow{10}{*}{Specialized Role} & 
Agent-001 is a \textcolor{xhsred}{error detection specialist} focused on identifying calculation mistakes, logical inconsistencies, and reasoning flaws in mathematical solutions.\\
\addlinespace
& Agent-002 is a \textcolor{xhsred}{logical reasoning specialist} specialized in mathematical proof construction, step-by-step deduction, and logical argument validation.\\
\addlinespace
& Agent-003 is a \textcolor{xhsred}{context comprehension specialist} responsible for understanding problem statements, extracting key information, and summarizing lengthy mathematical contexts.\\
\addlinespace
& Agent-004 is a \textcolor{xhsred}{computational specialist} dedicated to performing accurate calculations, numerical analysis, and algebraic manipulations.\\
\addlinespace
& Agent-005 is a \textcolor{xhsred}{solution verification specialist} focused on checking final answers, validating solution paths, and ensuring mathematical correctness.\\
\bottomrule
\end{tabular}
\end{table*}

\begin{figure*}[t]
\centering
\begin{minipage}[t]{0.49\textwidth}
\centering
\begin{tikzpicture}
\begin{axis}[
    width=6.5cm,
    height=5.5cm,
    colorbar,
    colormap/viridis,
    point meta min=28,
    point meta max=58,
    xlabel={Dialogue Rounds},
    ylabel={Number of Agents},
    xtick={0.5,1.5,2.5,3.5},
    ytick={0.5,1.5,2.5,3.5},
    xticklabels={2,3,4,5},
    yticklabels={2,3,4,5},
    nodes near coords,
    every node near coord/.append style={font=\tiny, text=white},
]
\addplot[
    matrix plot*,
    mesh/cols=4,
    point meta=explicit,
] table[meta=value] {
    x   y   value
    0   0   36
    1   0   35
    2   0   34
    3   0   33
    0   1   35
    1   1   34
    2   1   33
    3   1   32
    0   2   34
    1   2   33
    2   2   32
    3   2   31
    0   3   33
    1   3   32
    2   3   31
    3   3   30
};
\end{axis}
\end{tikzpicture}
\caption{\method w/ General Role performance on MATH-100}
\label{fig:heatmap_general}
\end{minipage}
\hfill
\begin{minipage}[t]{0.49\textwidth}
\centering
\begin{tikzpicture}
\begin{axis}[
    width=6.5cm,
    height=5.5cm,
    colorbar,
    colormap/viridis,
    point meta min=28,
    point meta max=58,
    xlabel={Dialogue Rounds},
    ylabel={Number of Agents},
    xtick={0.5,1.5,2.5,3.5},
    ytick={0.5,1.5,2.5,3.5},
    xticklabels={2,3,4,5},
    yticklabels={2,3,4,5},
    nodes near coords,
    every node near coord/.append style={font=\tiny, text=white},
    visualization depends on=\thisrow{show} \as \show,
    nodes near coords style={opacity=\show},
]
\addplot[
    matrix plot*,
    mesh/cols=4,
    point meta=explicit,
] table[meta=value] {
    x   y   value  show
    0   0   32     1
    1   0   34     1
    2   0   33     1
    3   0   31     1
    0   1   38     1
    1   1   41     1
    2   1   39     1
    3   1   36     1
    0   2   45     1
    1   2   48     1
    2   2   46     1
    3   2   42     1
    0   3   52     1
    1   3   58     0
    2   3   55     1
    3   3   49     1
};
\node at (axis cs:1,3.2) {\textbf{\textcolor{black}{58}}};
\end{axis}
\end{tikzpicture}
\caption{\method w/ Specified Role performance on MATH-100}
\label{fig:heatmap_specified}
\end{minipage}

\vspace{1em}

\begin{minipage}[t]{0.49\textwidth}
\centering
\begin{tikzpicture}
\begin{axis}[
    width=6.5cm,
    height=5.5cm,
    colorbar,
    colormap/viridis,
    point meta min=28,
    point meta max=58,
    xlabel={Dialogue Rounds},
    ylabel={Number of Agents},
    xtick={0.5,1.5,2.5,3.5},
    ytick={0.5,1.5,2.5,3.5},
    xticklabels={2,3,4,5},
    yticklabels={2,3,4,5},
    nodes near coords,
    every node near coord/.append style={font=\tiny, text=white},
]
\addplot[
    matrix plot*,
    mesh/cols=4,
    point meta=explicit,
] table[meta=value] {
    x   y   value
    0   0   42
    1   0   40
    2   0   39
    3   0   37
    0   1   50
    1   1   44
    2   1   43
    3   1   41
    0   2   45
    1   2   42
    2   2   40
    3   2   38
    0   3   43
    1   3   40
    2   3   38
    3   3   36
};
\end{axis}
\end{tikzpicture}
\caption{AutoGen performance on MATH-100}
\label{fig:heatmap_autogen}
\end{minipage}
\hfill
\begin{minipage}[t]{0.49\textwidth}
\centering
\begin{tikzpicture}
\begin{axis}[
    width=6.5cm,
    height=5.5cm,
    colorbar,
    colormap/viridis,
    point meta min=28,
    point meta max=58,
    xlabel={Dialogue Rounds},
    ylabel={Number of Agents},
    xtick={0.5,1.5,2.5,3.5},
    ytick={0.5,1.5,2.5,3.5},
    xticklabels={2,3,4,5},
    yticklabels={2,3,4,5},
    nodes near coords,
    every node near coord/.append style={font=\tiny, text=white},
]
\addplot[
    matrix plot*,
    mesh/cols=4,
    point meta=explicit,
] table[meta=value] {
    x   y   value
    0   0   44
    1   0   49
    2   0   46
    3   0   43
    0   1   42
    1   1   45
    2   1   44
    3   1   41
    0   2   40
    1   2   43
    2   2   42
    3   2   39
    0   3   38
    1   3   41
    2   3   40
    3   3   37
};
\end{axis}
\end{tikzpicture}
\caption{Multi-Agent Debate performance on MATH-100}
\label{fig:heatmap_debate}
\end{minipage}
\end{figure*}

\subsection{Overall Performance}

\subsubsection{Mathematical Reasoning Analysis}

\par\vspace{2\baselineskip}\noindent \begin{finding}{1}
\method\ achieves superior mathematical reasoning performance through effective task decomposition and agent-level specialization, with the greatest advantages manifesting in high-complexity problems where structured collaborative analysis outperforms single-agent approaches.
\end{finding}

Table~\ref{tab:math} demonstrates \method's consistent superiority across mathematical reasoning benchmarks. On GSM8K, \method\ achieves 87.41\% and 91.50\% accuracy with Qwen2.5-72B and Llama-3.1-70B respectively. The performance gap becomes more pronounced on challenging tasks: on AIME(2024), \method\ reaches 30.4\% accuracy with Qwen2.5-72B, nearly doubling other methods' performance. The substantial improvement from Naive (37.52\%) to Naive-CoT (75.13\%) on GSM8K highlights the importance of explicit reasoning, while ReAct's underperformance suggests its structured reasoning-action cycles may not align with pure mathematical reasoning demands.

\subsubsection{Code Generation Analysis}

\par\vspace{2\baselineskip}\noindent \begin{finding}{2}
\method\ excels at generating high-quality initial solutions through collaborative analysis, achieving superior pass@1 performance, though structured frameworks like ReAct demonstrate greater effectiveness in diversified sampling scenarios, indicating task-specific architectural requirements.
\end{finding}

Code generation results reveal \method's distinct performance characteristics. \method\ achieves optimal pass@1 scores: 79.20\% on HumanEval with Llama-3.1-70B and 76.46\% with Qwen2.5-72B. However, performance gains diminish at higher sampling rates (pass@5: 80.91\% and 84.15\%), contrasting with ReAct's improvement from 78.53\% to 87.66\%. This pattern suggests \method's collaborative mechanism rapidly produces high-quality solutions through multi-perspective analysis, but may limit exploration space in iterative optimization scenarios where diverse solution generation is crucial.

\subsubsection{Commonsense Reasoning Analysis}

\par\vspace{2\baselineskip}\noindent \begin{finding}{3}
\method's task decomposition and multi-agent coordination create unnecessary overhead for commonsense reasoning, where direct model responses and simple chain-of-thought prove more effective than collaborative analysis.
\end{finding}

Commonsense reasoning exposes \method's limitations when collaborative complexity exceeds task requirements. On HellaSwag, Naive methods achieve superior performance (73.7\% vs \method's 70.3\% on Qwen2.5-72B) because commonsense questions often have obvious correct answers that models can identify directly. \method's divide-and-conquer approach fragments these straightforward questions into unnecessary subtasks, while Naive-CoT's success (85.5\% on WinoGrande with Qwen2.5-72B) demonstrates that single-model reasoning with explicit thought processes suffices.

Multiple agents discussing obvious answers can introduce doubt and second-guessing, leading to incorrect consensus where initial correct interpretations may be overturned through excessive deliberation. AutoGen's consistent underperformance (64.4\% and 78.4\% on Qwen2.5-72B) further confirms that structured multi-agent workflows become counterproductive when applied to tasks where immediate model knowledge access is most effective.

\subsubsection{Structural Text Understanding Analysis}

\par\vspace{2\baselineskip}\noindent \begin{finding}{4}
\method\ demonstrates robustness in handling complex structured information through effective task decomposition, maintaining performance stability as context complexity increases, while reasoning-intensive methods exhibit significant degradation under structured data challenges.
\end{finding}

\method\ achieves optimal performance on high-complexity configurations, reaching 52.1\% accuracy at Width=3, Depth=3, significantly outperforming other methods. As complexity increases, reasoning-intensive approaches show marked degradation: Multi-Agent Debate drops from 83.3\% to 40.3\%, and Naive-CoT from 86.7\% to 42.3\%. The direct Naive method maintains relative stability (45.0\% on difficult configurations), while ReAct exhibits extreme performance collapse (below 1.2\%). These results indicate that \method's divide-and-conquer strategy effectively manages structured data's context explosion challenges.

\subsubsection{Domain-Specific Knowledge Analysis}

\par\vspace{2\baselineskip}\noindent \begin{finding}{5}
\method\ demonstrates domain-adaptive effectiveness, excelling in tasks requiring multi-perspective analysis (financial reasoning) while maintaining competitive performance across diverse professional domains, though complex interaction frameworks may introduce unnecessary overhead in knowledge-retrieval-intensive tasks.
\end{finding}

Domain-specific results reveal \method's balanced performance across professional areas. In finance (FinQual), collaborative approaches achieve superior results with Multi-Agent Debate reaching 80.20\%. Medical domain (MedmcQA) shows strong model dependency, with Llama-3.1-70B achieving 90.20\% via Multi-Agent Debate. Legal reasoning (JEC-QA) challenges all methods, with maximum accuracy of 42.56\%. \method\ maintains consistent performance across domains, while ReAct consistently underperforms, suggesting domain-specific architectural requirements for optimal effectiveness.

\subsection{Ablation Study}

To thoroughly understand the impact of different components in multi-agent systems on mathematical reasoning performance, we conducted comprehensive ablation experiments on the MATH-100 dataset. MATH-100 consists of 100 problems randomly sampled from Level 5 difficulty of the MATH dataset, specifically designed for parameter sensitivity analysis.

The experiments systematically evaluate the effects of two key dimensions—number of agents (2-5) and dialogue rounds (2-5)—on the performance of four different multi-agent frameworks. Additionally, we compare two distinct role configuration strategies in \method: general role configuration and specialized role configuration. As shown in Table~\ref{tab:abla_role}, the general role configuration assigns all agents the same mathematical expert identity, while the specialized role configuration implements fine-grained division of labor based on the cognitive process of mathematical problem solving. All experiments use Qwen2.5-72B as the underlying language model.

\subsubsection{Overall Performance Comparison}

Heat map analysis from Figure~\ref{fig:heatmap_general} to Figure~\ref{fig:heatmap_debate} reveals significant performance differences among the four methods. \method w/ Specified Role demonstrates the greatest performance potential across all configurations, achieving a maximum accuracy of 58\% (5 agents, 3 dialogue rounds), while \method w/ General Role reaches only 36\% (2 agents, 2 dialogue rounds). AutoGen achieves its best performance of 50\% (3 agents, 2 dialogue rounds), and Multi-Agent Debate reaches 49\% (2 agents, 3 dialogue rounds).

\par\vspace{2\baselineskip}\noindent \begin{finding}{7}
Specialized role division and homogeneous agent configuration exhibit completely opposite performance trends when the number of agents increases: the former improves with more agents while the latter deteriorates.
\end{finding}

\method w/ Specified Role shows a strong positive correlation: accuracy significantly improves from an average of 32.5\% with 2 agents to 53.5\% with 5 agents, representing a 64.6\% improvement. Each additional specialized agent contributes approximately 7 percentage points of performance gain. This consistent growth pattern indicates that specialized division of labor can avoid information redundancy through collaboration among different roles.

Conversely, \method w/ General Role exhibits a continuous performance decline as the number of agents increases: from an average accuracy of 34.5\% with 2 agents to 31.5\% with 5 agents, a decrease of 8.7\%. When all agents assume the same mathematical expert role, adding more agents generates similar viewpoints and redundant information, leading to ineffective discussions rather than productive collaboration.

\par\vspace{2\baselineskip}\noindent \begin{finding}{8}
\method can effectively coordinate large-scale agent teams for collaboration, while traditional multi-agent frameworks suffer performance degradation as agent numbers increase, failing to achieve effective large-scale cooperation.
\end{finding}

Both AutoGen and Multi-Agent Debate show performance decline with increasing agent numbers. AutoGen reaches its best average performance of 44.5\% with 3 agents, then drops to 39.25\% with 5 agents. Multi-Agent Debate achieves its highest average performance of 45.5\% with 2 agents, then continuously declines to 39\% with 5 agents. This suggests that traditional multi-agent frameworks cannot effectively manage interactions among more agents, resulting in reduced collaboration efficiency as scale increases.

In contrast, \method w/ Specified Role not only performs best in large-scale configurations but also exhibits the greatest performance potential range: 87.1\% difference between optimal and worst configurations (58\% vs 31\%). Other methods show relatively limited performance ranges: AutoGen at 38.9\%, Multi-Agent Debate at 35.1\%, and \method w/ General Role at only 28.6\%. This demonstrates that \method achieves more effective large-scale agent collaboration through specialized role division.

\par\vspace{2\baselineskip}\noindent \begin{finding}{9}
Specialized role division requires moderate interaction depth to integrate diverse perspectives, while homogeneous agents cannot achieve performance improvements by increasing dialogue rounds.
\end{finding}

\method w/ Specified Role exhibits a pattern of initial increase followed by decrease: across most agent configurations, moderate interaction depth achieves optimal performance. For example, with 5 agents, accuracy rises from 52\% at 2 rounds to 58\% at 3 rounds, then drops to 49\% at 5 rounds. This indicates that different specialized roles require sufficient interaction to integrate their professional insights, but excessive interaction can lead to decision complexity.

Conversely, \method w/ General Role is relatively insensitive to dialogue round variations, with accuracy slightly declining from 34.5\% at 2 rounds to 31.5\% at 5 rounds, a change of approximately 8.7\%. Since all agents share the same professional background, extended dialogue mainly produces repetitive information and cannot gain new cognitive perspectives or quality improvements through increased interaction depth.

AutoGen performs best at 2 dialogue rounds, while Multi-Agent Debate reaches optimum at 3 rounds, both showing performance decline with further round increases, indicating that different collaboration mechanisms have their respective optimal interaction depth configurations.

\section{Case Study}
\begin{table}[!ht]
\centering
\caption{Agent Assignment for Interactive Data Visualization Tool Development Tasks}
\label{tab:visualization_agents}
\small
\begin{tabular}{|m{3cm}|m{4.5cm}|}
\hline
\textbf{Task Name} & \textbf{Participating Agents} \\ \hline
Module Interface Design & Requirement Analyst Agent, Code Implementation Agent, Code Review Agent \\ \hline
File Parsing Implementation & Requirement Analyst Agent, Code Implementation Agent, Code Review Agent \\ \hline
Data Processing Implementation & Requirement Analyst Agent, Code Implementation Agent, Code Review Agent \\ \hline
Data Display Implementation & Requirement Analyst Agent, Code Implementation Agent, Code Review Agent \\ \hline
Interactive Function Implementation & Requirement Analyst Agent, Code Implementation Agent, Code Review Agent \\ \hline
Testing and Verification & Test Planning Agent, Test Execution Agent, Quality Assurance Agent \\ \hline
\end{tabular}
\end{table}

\begin{table}[t]
\centering
\caption{Agent Assignment for Blockchain Technology Analysis Article Writing Tasks}
\label{tab:article_agents}
\small
\begin{tabular}{|m{3cm}|m{4.5cm}|}
\hline
\textbf{Task Name} & \textbf{Participating Agents} \\ \hline
Technology Survey & Research Planning Agent, Research Execution Agent, Content Review Agent \\ \hline
Case Collection & Research Planning Agent, Research Execution Agent, Content Review Agent \\ \hline
Market Analysis & Research Planning Agent, Research Execution Agent, Content Review Agent \\ \hline
Writing Technology Analysis Chapters & Writing Planning Agent, Writing Execution Agent, Content Review Agent \\ \hline
Writing Market Analysis Chapters & Writing Planning Agent, Writing Execution Agent, Content Review Agent \\ \hline
Overall Optimization & Integration Planning Agent, Integration Execution Agent, Quality Assurance Agent \\ \hline
\end{tabular}
\end{table}

\begin{figure}[!ht]
\centering
\includegraphics[width=\linewidth]{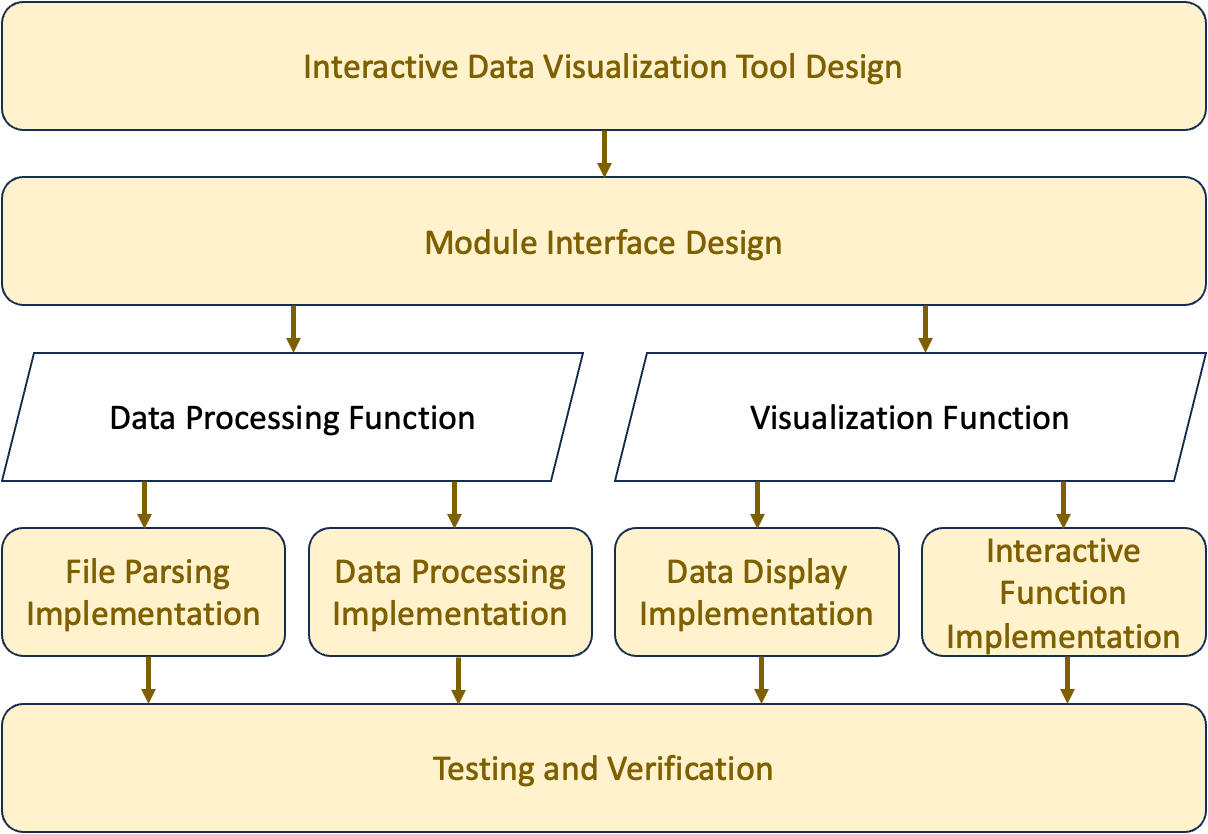}
\caption{A schematic illustration of task decomposition for the ``Interactive Data Visualization Tool Development'' task by \method.
Rectangular boxes represent specific executable tasks, while parallelograms denote annotation labels for task groupings.
Arrow connections indicate dependency relationships between tasks, and all tasks not directly connected can be executed in parallel after the completion of their prerequisite tasks.}
\label{fig:case1}
\end{figure}

\begin{figure}[!ht]
\centering
\includegraphics[width=0.6\linewidth]{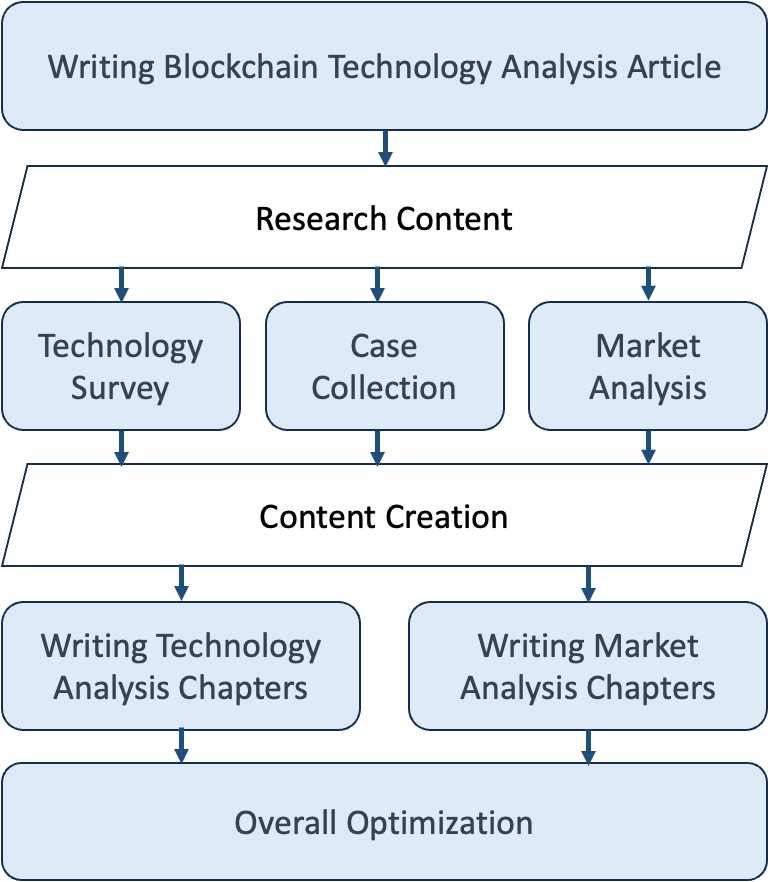}
\caption{Another illustration of task decomposition for ``Writing Blockchain Techonology Analysis Article'' task by \method}
\label{fig:case2}
\end{figure}

\begin{table*}[!ht]
\centering
\caption{Two-Round Agent Dialogue for File Parsing Task}
\label{tab:file_parsing_dialogue}
\small
\begin{tabular}{|c|m{3.5cm}|m{12cm}|}
\hline
\textbf{Round} & \textbf{Agent} & \textbf{Dialogue Content} \\ \hline
\multirow{12}{*}{\rotatebox{90}{\textbf{Round 1}}} 
& Requirement Analyst Agent & Let me analyze this file parsing task... We need to implement a parsing function supporting CSV and Excel files. Core requirements include: 1) File format validation; 2) Automatic character encoding detection; 3) Data parsing with JSON output; 4) Large file handling capability... I suggest implementing the basic framework first. \\ \cline{2-3}

& Code Implementation Agent & \begin{lstlisting}[language=Python, basicstyle=\tiny]import pandas as pd
def parse_file(file_path):
    try:
        if file_path.endswith('.csv'):
            df = pd.read_csv(file_path)
        elif file_path.endswith('.xlsx'):
            df = pd.read_excel(file_path)
        else:
            return {"error": "Unsupported format"}
        # ... data processing logic
        return df.to_dict('records')
    except Exception as e:
        return {"error": str(e)}\end{lstlisting}\\ \cline{2-3}

& Code Review Agent & Reviewing the implementation... The basic functionality exists, but several issues need attention: 1) No file size limitation, potential memory overflow; 2) Encoding detection relies entirely on pandas defaults, may cause garbled text; 3) Exception handling too simplistic, poor user experience; 4) Missing data type inference capability... Requires substantial improvements. \\ \hline

\multirow{15}{*}{\rotatebox{90}{\textbf{Round 2}}}
& Requirement Analyst Agent & Based on the code review feedback... Adding specific requirements: 1) 100MB file size limit; 2) Support UTF-8, GBK, ASCII encoding auto-detection; 3) Categorized error handling with user-friendly messages; 4) Intelligent data type recognition for dates, numbers, and text; 5) Chunked processing mechanism for large files... This should provide comprehensive coverage.\\ \cline{2-3}

& Code Implementation Agent & \begin{lstlisting}[language=Python, basicstyle=\tiny]import pandas as pd
import chardet
import os
from typing import Callable, Dict, Any

class FileParser:
    def __init__(self, chunk_size=10000):
        self.chunk_size = chunk_size
    
    def parse_file(self, file_path: str) -> Dict[str, Any]:
        # File size validation
        if os.path.getsize(file_path) > 100*1024*1024:
            return {"error": "File exceeds 100MB limit"}
        
        # Enhanced encoding detection
        with open(file_path, 'rb') as f:
            raw_data = f.read(50000)
            detected = chardet.detect(raw_data)
            encoding = detected['encoding'] if detected['confidence'] > 0.7 else 'utf-8'
        
        try:
            if file_path.endswith('.csv'):
                # ... chunked CSV processing
                return self._parse_csv_chunked(file_path, encoding)
            elif file_path.endswith(('.xlsx', '.xls')):
                # ... Excel processing with compatibility
                return self._parse_excel(file_path)
        except Exception as e:
            return {"error": f"Parse error: {str(e)}", "type": type(e).__name__}
    
    def _parse_csv_chunked(self, file_path, encoding):
        # ... streaming processing implementation
        chunks = []
        for chunk in pd.read_csv(file_path, encoding=encoding, chunksize=self.chunk_size):
            chunks.append(chunk)
        df = pd.concat(chunks, ignore_index=True)
        return {"data": df.to_dict('records'), "dtypes": df.dtypes.to_dict()}\end{lstlisting}\\ \cline{2-3}

& Code Review Agent & Final review assessment...  \\ \hline
\end{tabular}
\end{table*}

\subsection{Task Tree}

To demonstrate AgentGroupChat's divide-and-conquer strategy, we present one representative scenarios illustrating how complex problems can be decomposed into sub-tasks and solved through specialized agent collaboration.

As illustrated in Figure \ref{fig:case1}, the interactive data visualization tool development is decomposed into four primary phases: module interface design, data processing functionality, visualization functionality, and testing verification. Upon completion of the module interface design, data processing and visualization functions can be developed in parallel, followed by unified testing verification. Each specific task is executed by specialized agents as listed in Table \ref{tab:visualization_agents}.

Figure \ref{fig:case2} demonstrates the workflow for composing a blockchain technology analysis article, which begins with three parallel research tasks: technology survey, case collection, and market analysis. Subsequently, based on the research findings, technical analysis and market analysis chapters are written in parallel, followed by overall optimization. As shown in Table \ref{tab:article_agents}, each phase is configured with corresponding specialized agents to ensure content quality and professionalism.

\subsection{Group Chat}

To provide a concrete demonstration of inter-agent collaboration patterns, we present a file parsing task as an exemplar, illustrating how three specialized agents progressively refine solutions through two rounds of dialogue.

Through these two rounds of authentic dialogue, the collaboration pattern among the three agents becomes evident: the Requirement Analyst Agent progressively transforms abstract requirements into technical specifications, the Code Implementation Agent iteratively develops solutions based on feedback, and the Code Review Agent ensures code quality from multiple perspectives.

\section{Conclusion}
This paper presents \method, a general-purpose framework for LLM-based multi-agent systems that systematically addresses critical challenges in the field through innovative architectural design and collaboration mechanisms. Our key contributions include a divide-and-conquer parallel architecture with hierarchical task forest structures, an adaptive collaboration engine for dynamic agent configuration, and optimized agent organization strategies—collectively enabling efficient distributed processing and superior performance across domains. Experimental results validate these contributions, with \method achieving substantial improvements on mathematical reasoning (91.50\% on GSM8K, 30.4\% on AIME), code generation (79.20\% pass@1 on HumanEval), and various reasoning tasks. Most notably, performance advantages consistently increase with task complexity, demonstrating that \method's approach effectively leverages collective agent intelligence to solve challenging problems that exceed single-agent capabilities. By combining generalizability with computational efficiency and performance guarantees, \method establishes a new paradigm for practical, high-performance multi-agent systems capable of addressing increasingly complex real-world applications.


\bibliography{neurips_2024}
\bibliographystyle{neurips_2024}





\end{CJK}
\end{document}